%% file: main.tex
\documentclass{article}

\usepackage[utf8]{inputenc}

\usepackage{microtype}
\usepackage{graphicx}
\usepackage{subcaption}
\usepackage{booktabs} 

\usepackage{hyperref}

\usepackage[preprint]{icml2026}

\usepackage{amsmath}
\usepackage{amssymb}
\usepackage{mathtools}
\usepackage{amsthm}

\usepackage{multirow}
\usepackage[fencedCode, hybrid]{markdown}
\usepackage{tablefootnote}
\usepackage{colortbl}
\usepackage{svg}
\usepackage{pgfplots}
\usepackage{tikz}
\pgfplotsset{compat=1.18}
\usepgfplotslibrary{colorbrewer}
\usetikzlibrary{pgfplots.fillbetween}

\usepackage{minted}
\usepackage{tcolorbox}
\tcbuselibrary{minted} 

\usepackage{xcolor}
\definecolor{codegray}{gray}{0.9} 

\newcommand{\stdfont}[1]{{\tiny #1}}

\usepackage[capitalize,noabbrev]{cleveref}

\theoremstyle{plain}

\theoremstyle{definition}

\theoremstyle{remark}

\usepackage[textsize=tiny]{todonotes}

\icmltitlerunning{Rogue One}

\begin{document}

\twocolumn[
    \icmltitle{Knowledge-Informed Automatic Feature Extraction\\ via Collaborative Large Language Model Agents}



  \icmlsetsymbol{equal}{*}

  \begin{icmlauthorlist}
    \icmlauthor{Henrik Brådland}{pitt,uia,norkart}
    \icmlauthor{Morten Goodwin}{uia}
    \icmlauthor{Vladimir I. Zadorozhny}{pitt,uia}
    \icmlauthor{Per-Arne Andersen}{uia}
  \end{icmlauthorlist}
    
  \icmlaffiliation{pitt}{School of Computing and Information, University of Pittsburgh, Pittsburgh, USA}
  \icmlaffiliation{uia}{Centre for Artificial Intelligence Research, University of Agder, Grimstad, Norway}
  \icmlaffiliation{norkart}{Norkart AS, Oslo, Norway}


  \icmlcorrespondingauthor{Henrik Brådland}{henrik.bradland@uia.no}

  \icmlkeywords{Machine Learning, ICML}

  \vskip 0.3in
]



\printAffiliationsAndNotice{}  

\input{sections/abstract}

\input{sections/introduction}
\input{sections/relatedwork}
\input{sections/method}
\input{sections/experiment}
\input{sections/discussion}

\input{sections/conclution}


\bibliography{references, references-2}
\bibliographystyle{icml2026}

\newpage
\appendix
\onecolumn

\newpage
\input{apendix/prompts}
\newpage
\input{apendix/test_assesments}

\newpage
\input{apendix/run_time}


\end{document}

%% file: sections/abstract.tex
\begin{abstract}

The performance of machine learning models on tabular data is critically dependent on high-quality feature engineering. While Large Language Models (LLMs) have shown promise in automating feature extraction (AutoFE), existing methods are often limited by monolithic LLM architectures, simplistic quantitative feedback, and a failure to systematically integrate external domain knowledge. This paper introduces Rogue One, a novel, LLM-based multi-agent framework for knowledge-informed automatic feature extraction. Rogue One operationalizes a decentralized system of three specialized agents—Scientist, Extractor, and Tester—that collaborate iteratively to discover, generate, and validate predictive features. Crucially, the framework moves beyond primitive accuracy scores by introducing a rich, qualitative feedback mechanism and a "flooding-pruning" strategy, allowing it to dynamically balance feature exploration and exploitation. By actively incorporating external knowledge via an integrated retrieval-augmented (RAG) system, Rogue One generates features that are not only statistically powerful but also semantically meaningful and interpretable. We demonstrate that Rogue One significantly outperforms state-of-the-art methods on a comprehensive suite of 19 classification and 9 regression datasets. Furthermore, we show qualitatively that the system surfaces novel, testable hypotheses, such as identifying a new potential biomarker in the myocardial dataset, underscoring its utility as a tool for scientific discovery.

\end{abstract}

%% file: sections/introduction.tex
\section{Introduction}
\label{sec:introduction}

The rising need for interpretable machine learning, combined with their documented challenges on smaller tabular datasets~\cite{grinsztajn_why_nodate}, has led to the continued dominance of classical tree-based models like Random Forest and XGBoost for tabular prediction. Nevertheless, the performance of the tree-based models is heavily tied to the predictive power of the underlying features. This introduces a shift from a model-centric development to a data-centric development, where synthesizing strong features is the driver for stronger downstream performance. Traditional approaches use the intuition of domain experts to generate strong features. Although effective, this method is highly subjective and does not well explore the space of possible features. Research has proposed to automate the process, thus giving rise to the field of Automatic Feature Extraction (AutoFE). Traditional AutoFE methods (e.g., polynomial feature generation) struggled to create semantically meaningful features, as they lacked the ability to incorporate domain knowledge or understand the high-level relationships between feature concepts. Nevertheless, with the rise of Large Language Models (LLMs), with their strong contextual understanding and parametric knowledge, they represent a powerful new paradigm for discovering new features as demonstrated by Abhyankar et. al.~\cite{abhyankar_llm-fe_2025} and Nam \& Kim~\cite{nam_optimized_2024}.

While promising, these methods treat the LLM as a monolithic generator, do not integrate diverse sources of learning signals, and rely solely on its internal parametric knowledge. To address this, we re-frame the problem around three key capabilities: (1) integrating qualitative learning signals beyond simple scalar metrics, (2) using a flexible Multi-Agent network instead of a monolithic generator, and (3) incorporating external knowledge to augment the LLM's parametric memory. With these three new capabilities, we present \textbf{Rogue One}, an AutoFE framework powered by modern Agentic LLM capabilities. The design forces the LLMs to provide explanations as intermediate steps. This, combined with the use of symbolic criteria for feature generation, results in interpretable features that potentially provide new insight into the underlying domain. Thus, Rogue One can be a tool to acquire insight into data in a human-interpretable fashion, thereby advancing other fields of research like medicine, finance, and engineering.

The Rogue One framework consists of iterating over a network of LLM-agents. The overall goal of the network is to perform \textbf{intelligent feature discovery}, implying that the system: (1) possesses an overall strategy for what kind of attributes to search for. (2) Can generalize and learn from previous iterations. (3) Can actively utilize external domain knowledge when applicable.

\paragraph*{Key contributions:}
\begin{itemize}
    \item We propose Rogue One, a novel LLM-based Multi-Agent framework for Automated Feature Extraction, built upon qualitative learning signals and flexible multi-agent network structures.
    \item We demonstrate that Rogue One outperforms the state-of-the-art methods over a variety of classification and regression tasks while simultaneously providing interpretability.
    \item We perform analysis to uncover the strengths and limitations of Rogue One.
\end{itemize}

\paragraph*{Paper outline}
The rest of the paper is structured as follows: Section~\ref{sec:related_work} provides a brief introduction to related work, while Section~\ref{sec:method} introduces the methodology and system design for Rogue One. Section~\ref{sec:results} describes the experimental setup, evaluation, and results, which are discussed in Section~\ref{sec:discussion} and concluded in Section~\ref{sec:conclusion}.

%% file: sections/relatedwork.tex
\section{Related work}
\label{sec:related_work}

Tabular models, such as XGBoost~\cite{Chen_2016} and HyperFast~\cite{article}, rely on strong features to perform well in various areas, including finance, medicine, and science. Traditionally, these strong features are handcrafted by domain experts, relying heavily on human intuition and domain expertise \cite{chandra_applications_2025}. Although effective, this hinders exploration by constraining the model to operate on a fixed feature space. The field of Automatic Feature Extraction (AutoFE) intends to overcome this by automatically generating features, be it temporal, spatial, or statistical. Systems like AutoFeat~\cite{10597758} and OpenFE~\cite{10.5555/3618408.3620168} provide a wide set of features to fit the original data, but lack a broader perspective of optimization, namely an optimization process that encompasses domain knowledge so as to make new features based on logic. Newer approaches~\cite{abhyankar_llm-fe_2025, nam_optimized_2024, hollmann2023largelanguagemodelsautomated,  han_large_2024, ouyang_fela_2025} propose to use an LLM, driven by its parametirc knowledge, to transform the raw data into new attributes. The setups rely on feedback loops where LLMs propose features, and a traditional ML model is trained whose characteristics are then used to improve the next iteration of feature extraction. Han et. al.~\cite{han_large_2024}, Hollmann et. al.~\cite{hollmann_caafe_nodate}, and Abhyankar et. al.~\cite{abhyankar_llm-fe_2025} provide the LLMs with a Python environment, thus utilizing the models' code-writing capabilities to express data transformations. All established approaches use a single LLM architecture~\cite{nam_optimized_2024, hollmann2023largelanguagemodelsautomated,   han_large_2024, abhyankar_llm-fe_2025}, thus not leveraging the potential in multi-agent collaboration~\cite{tran_multi-agent_2025}. Furthermore, Abhyankar et. al.~\cite{abhyankar_llm-fe_2025} show the importance of domain knowledge in their ablation study, yet they and most others rely fully on the paramedic knowledge of the LLM and In-Context Examples. On the contrary, modern methods of incorporating external knowledge, like Retrial-Augmented Generation (RAG), allow models to substantially extend their knowledge. Lastly, all earlier literature~\cite{nam_optimized_2024, hollmann2023largelanguagemodelsautomated, han_large_2024, abhyankar_llm-fe_2025, ouyang_fela_2025} relies on simplistic, quantitative rewards (e.g., model accuracy) as the sole feedback signal, whereas Rogue One introduces a rich, qualitative assessment of feature quality, stability, and redundancy, which LLM-Agents are shown to be capable of producing~\cite{zheng_automation_2025}. 

Ouyang and Wang~\cite{ouyang_fela_2025} very recently proposed FELA, the first Multi-Agent AutoFE architecture. Although addressing the use of external knowledge and collaborative agents, it still relies on quantitative rewards, a rigid information flow, and an unnecessary level of complexity. FELA was not included in our quantitative comparison due to the lack of a publicly available implementation and reporting of scores on established datasets. Our work differs from FELA's by employing a more flexible agentic design and a richer, qualitative feedback mechanism, as opposed to FELA's reliance on quantitative rewards.



%% file: sections/method.tex
\section{Method}
\label{sec:method}

The feature extraction task is formulated as an optimization problem where the goal is to discover the set of optimal features $\mathcal{X}^* \in \mathbb{R}^{k \times n}$ consisting of $n$ feature vectors $\Vec{x^*} \in \mathbb{R}^{k}$. Each element of $\Vec{x^*}$ is the result of applying the transformations $h: \mathbb{R}^{m} \rightarrow \mathbb{R} $ on the corresponding row-vector $\Vec{x} \in \mathbb{R}^m$ from the raw data $\mathcal{X} \in \mathbb{R}^{m \times n}$. We denote $h(\mathcal{X})$ as applying $h(\cdot)$ to all $n$ row-vectors in $\mathcal{X}$, thus resulting in a column vector of optimal features. To simplify, we use the following notation when applying all learned transformations:

\begin{equation}
    \mathcal{X}^* = \mathcal{H}(\mathcal{X}) = \{h_1(\mathcal{X}), h_2(\mathcal{X}), \dots, h_{m^*}(\mathcal{X}) \}
\end{equation}

The optimization task can therefore be framed as: 

\begin{equation}
    \max_{\mathcal{H}} \mathcal{E}(f^*(\mathcal{H}(\mathcal{X}_{\text{val}})), \mathcal{Y}_{\text{val}}) \label{eq:optimization_goal}
\end{equation}
\begin{center}
subject to:
\end{center}
\begin{equation}
    f^* = \arg \min_{f} \mathcal{L}_f(f(\mathcal{H}(\mathcal{X}_{\text{train}})), \mathcal{Y}_{\text{train}})
\end{equation}

Where $\mathcal{E}$ is the evaluation metric (e.g., accuracy, recall, or RMSE) and $f^*$ is a predictor model with optimal parameters (e.g., Random Forest or XGBoost). $\mathcal{Y}$ is the target variable (e.g., one-hot encoded labels for classification or numerical values for regression), and $\mathcal{L}$ denotes the cost function that $f^*$ is optimized for. In other words, Rogue One opts to find the set of transformations $\mathcal{H}$ that $\mathcal{E}$ when applied to a known tabular dataset $\mathcal{X}$. For instance, given raw data $\mathcal{X}$ with columns \texttt{age}, \texttt{blood\_pressure}, and \texttt{height}, a transformation $h_1$ from the set $\mathcal{H}$ could be the symbolic function $h_1(\vec{x}) = \vec{x}_{\text{age}} \times \vec{x}_{\text{blood\_pressure}}$, while another transformation $h_2$ might be $h_2(\vec{x}) = \vec{x}_{\text{height}} / \vec{x}_{\text{age}}$. 




\subsection{System Overview}

\begin{figure*}[ht]
    \centering
    \includegraphics[width=0.99\linewidth]{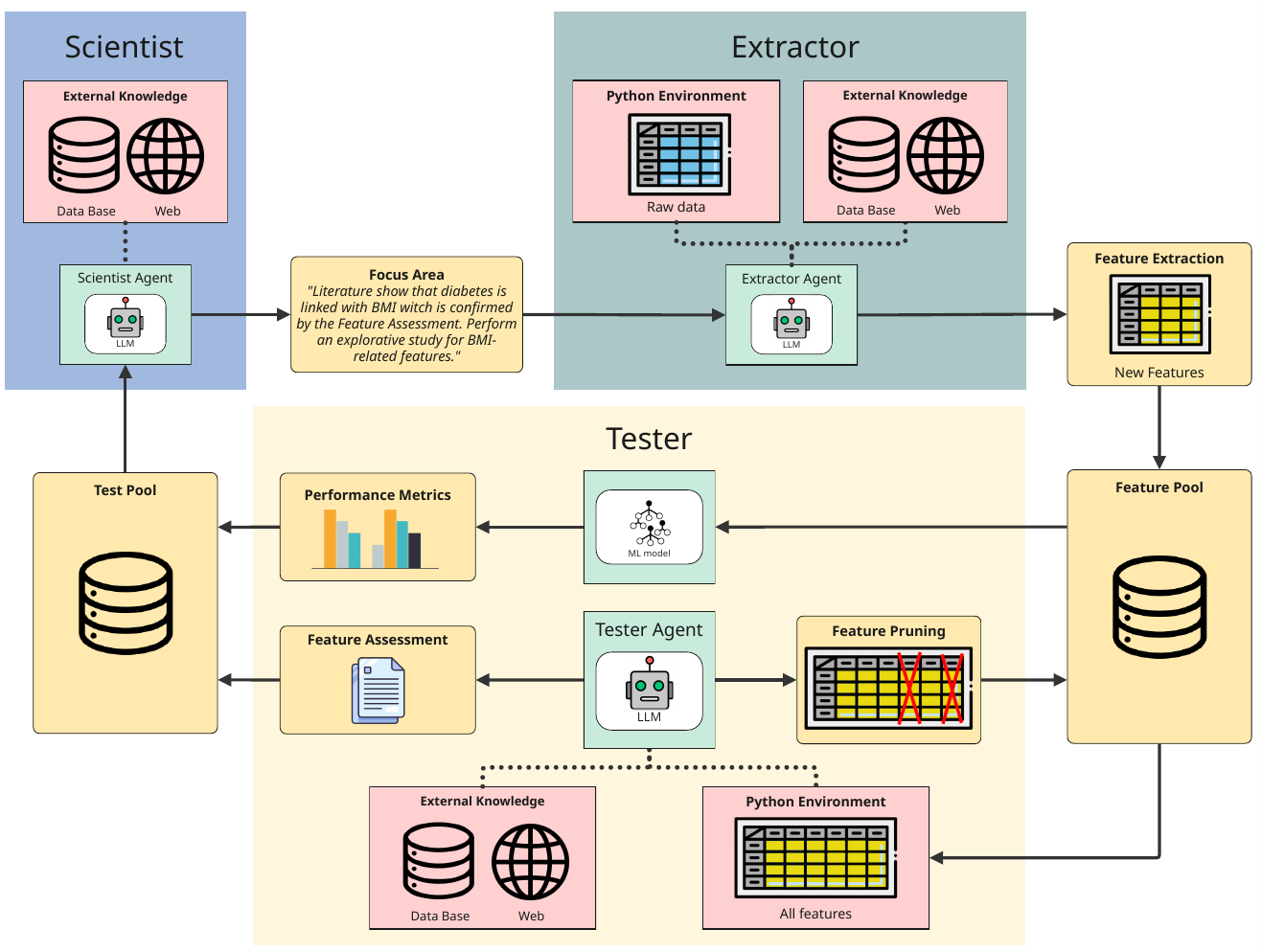} 
    \caption{The Rogue One system architecture operates as a continuous iterative loop, organized into three core stages, each managed by a specialized agent. The cycle commences with the Scientist Agent (blue background), which analyzes the Test Pool data from prior iterations to define a new Focus Area. Subsequently, the Extractor Agent (green) leverages this Focus Area to generate new candidate features, which are then appended to the central Feature Pool. Finally, the Tester Agent (yellow) orchestrates the evaluation and pruning phase. It employs ML models to generate Performance Metrics and a Feature Assessment, which together update the Test Pool. Concurrently, its Feature Pruning function refines the Feature Pool, completing the loop and preparing the system for the next iteration.
    }
    \label{fig:system_architecture}
\end{figure*}

The framework consists of three agents—the Scientist Agent, the Extractor Agent, and the Tester Agent —operating as a decentralized role-based Multi-Agent system\cite{tran_multi-agent_2025}, as illustrated in Figure~\ref{fig:system_architecture}. All three agents of Rogue One are designed to explore the raw data in an open-ended manner, in contrast to FELA~\cite{ouyang_fela_2025}, which limits the exploration-tasked Agent to only the data schema, thereby not accounting for statistical properties within raw data. Also, Rogue One implements a flooding strategy where tens of attributes are created each iteration (17 features on average during the experiments from Section~\ref{sec:results}), for then being pruned based on statistical testing by the Tester Agent, in contrast to other AutoFE systems~\cite{ouyang_fela_2025, hollmann_caafe_nodate, abhyankar_llm-fe_2025} which focus on extracting only a few high-quality features. There are no strict pruning rules; however, the Tester Agent is prompted to prune features that do not contribute meaningfully to the prediction task in terms of predictive power, redundancy, and robustness.

The operational cycle proceeds as follows for the $i$-th iteration:
\begin{enumerate}
    \item The \textbf{Scientist Agent} analyzes the cumulative results from all $i-1$ previous cycles, stored in a central \textbf{Test Pool}, to formulate a high-level search strategy. This strategy is articulated as a natural language directive called the \textbf{Focus Area}.
    
    \item The \textbf{Extractor Agent} receives the Focus Area and generates new candidate features by synthesizing and executing Python code to express $\mathcal{H}_i$. This process produces the numerical matrix $\mathcal{X}_i^*$ of new features and a structured JSON file of \textbf{Feature Explanations} containing the generating code and a semantic description for each feature. The new features and explanations are appended to the \textbf{Feature Pool}, which stores features and explanations from previous iterations. The feature matrix $\mathcal{X}^*$ is produced by concatenating extracted features from all iterations $\mathcal{X}^* = \{\mathcal{X}_1^*; \dots; \mathcal{X}_i^* \}$. 
    
    \item Equipped with a Python environment, the \textbf{Tester Agent} plans and executes tests to assess the quality of $\mathcal{X}^*$, indirectly also assessing $\mathcal{H}$. In doing so, the Agent prunes redundant and low-impact features and generates the \textbf{Feature Assessment}. The predictor model $f^*$ is trained and evaluated on the pruned data to calculate evaluation metrics using $\mathcal{E}(f^*)$. The feature assessment and evaluation metrics are stored in the \textbf{Test Pool} used by the Scientist Agent for the upcoming iterations.
   
\end{enumerate}

A knowledge access tool, based on a Single Agentic Retrieval-Augmented Generation (RAG) setup \cite{singh_agentic_2025}, is made available to all the Agents to allow access to external knowledge. The RAG setup consists of an LLM agent tasked to decompose the user query into sub-queries, perform retrieval for each sub-query, and curate a summary that is passed back to the user. The choice of external knowledge source is a design choice for any implementation, but it helps the aforementioned agents by expanding their expertise beyond their parametric knowledge. The source is a curated database of domain literature for complex research-intensive or sensitive domains, like medicine, or an open-ended web search for general domains. 

\subsection{Scientist Agent}
\label{sub:scientist_agent}

The Scientist Agent orchestrates the feature discovery process by setting the Focus Area for each iteration. Its primary function is to intelligently guide the search towards promising regions of the feature space by evaluating and assessing test results together with domain knowledge, thus learning from previous mistakes and successes, and utilizing prior knowledge to explain the feature relations.

The role of the Scientist Agent is to guide the long-term strategies for the agent-network, balancing exploration vs exploitation, and ensuring a knowledge-informed feature exploration. The Scientist Agent, therefore, takes the role of orchestration within the network, similar to a project manager. The agent starts assessing the results from previous cycles stored in the Test Pool, combined with the definitions for the current features. Using its reasoning capabilities, the LLM analyzes these top-performing features to identify a “common ground”—an underlying semantic or structural pattern (e.g., features derived from vital signs within the first 12 hours). It may query the knowledge access tool to enrich this analysis with external domain knowledge. Based on this analysis, the agent determines the search scope:
\begin{itemize}
    \item \textit{Exploitive scope:} If a clear, promising pattern is identified, the Focus Area directs the Extractor Agent to generate variations and more complex combinations related to the particular pattern.
    \item \textit{Exploratory scope:} If no strong pattern emerges or if a particular data modality is underexplored, the Focus Area directs the Extractor Agent to investigate a new area of the data.
\end{itemize}
The agent is equipped with a scratch pad, which it is instructed to use extensively to avoid having important details being lost within the model's context. When done, the model revisits its notes and forms the \textbf{Focus Area}, a text string that serves as the instruction for the next cycle.

\subsection{Extractor Agent}
\label{sub:extractor_agent}

The Extractor Agent is responsible for translating the strategic Focus Area into concrete, machine-executable code representing symbolic combinations of the raw data. It therefore takes on the role as the main workforce, not concerned with long-term planning, but focused on doing an isolated task, namely the extraction work. It does so by first undergoing a discovery phase, leveraging a Python environment to explore the raw feature set $\mathcal{X}$ (from simple commands like \texttt{df.head()} and \texttt{df.describe()}, to complex multi-line queries) to load relevant data schemas and statistical properties into its context window. Equipped with a scratch pad, it notes down useful insights from the discovery phase for further use. In the second phase, known as the extraction phase, the LLM provides pairs of a transformation $h$ and corresponding justifications for why $h(\mathcal{X})$ is rooted in the focus area. Requiring a justification for the transformation forces the model not to suggest arbitrary features, but rather features that align with the focus area. When the agent finishes its work, it has produced the set of transformations $\mathcal{H}_i$ and the corresponding symbolic features $\mathcal{X}_i^* = \mathcal{H}_i(\mathcal{X})$, stored in the Feature Pool, which are passed to the Tester Agent.

\subsection{Tester Agent}
\label{sub:tester_agent}

The Tester is a two-component module; the first being an evaluation of the predictor model $f^*$, and the second being an assessment of the features $\mathcal{X}^*$ carried out by the Tester Agent. The Tester is tasked with being the critical thinker, evaluating the result of the work carried out by the two other Agents. It therefore takes the role of a lab assistant or similar to a "red team" from software development. 

The evaluation performs a 5-fold validation using an XGBoost model as $f$. It then reports the performance metrics produced by $\mathcal{E}$, which serves as the global performance indicator for $\mathcal{X}^*$. The Tester Agent, equipped with a general Python environment, inspects $\mathcal{X}^*$ and the corresponding $\mathcal{H}$ to get an overview of the kind of features to assess (temporal, geometric, spatial,...). It then uses the external knowledge source to find the most appropriate way of assessing the features with respect to feature importance (predictive power), stability (resilience to noise), inter-feature relation (redundancy and correlations), and any other evaluation criterion it deems necessary. This allows for a rich, detailed, and human-readable feedback signal, in contrast to a primitive signal set by previous methods like LLM-FE~\cite{abhyankar_llm-fe_2025} and FELA~\cite{ouyang_fela_2025}. As a part of the testing, the Tester Agent also prunes $\mathcal{X}^*$, hence discarding features deemed redundant or with low predictive power. This is an essential counter mechanism to the flooding strategy of the Extractor Agent to reduce the feature count and to overcome the curse of dimensionality. During its work, the agent notes down its findings in the scratch pad, similar to the other Agents. When finished, the agent produces a comprehensive markdown document, known as the \textbf{Feature Assessment} (See example in Appendix~\ref{ape:test_assessments}), where it structures its findings. The assessment does not serve as a recommendation for further investigation, but rather an overview of the current feature set. This is done to keep a clear division of tasks between the Tester Agent and the Scientist Agent. The performance metrics and the feature assessment are passed on and appended as a new entry to the \textbf{Test Pool}, completing the cycle.

%% file: sections/experiment.tex
\section{Experiment setup and results}
\label{sec:results}

Rogue One is implemented using the \texttt{GPT-OSS-120B}\footnote{\url{https://huggingface.co/openai/gpt-oss-120b}} model\cite{openai_gpt-oss-120b_2025} is used as the LLM for all Agents without any fine-tuning. The prompting follows the role-based collaboration strategy, allowing for explicit labor division and leveraging models' own expertise. All system prompts are listed in Appendix~\ref{ape:prompt}. The knowledge sources used by the lookup agent, as shown in Figure 3, consist of relevant literature for the Extractor and Tester Agents, accessed through dense vector retrieval using the \texttt{Qwen3-Embedding-4B}\footnote{\url{https://huggingface.co/Qwen/Qwen3-Embedding-4B}} embedding model~\cite{qwen3embedding}. For the Scientist Agent, the vector database is swapped for a web search to account for the variety of domains within the test data. Rogue One was run for 10 iterations on all datasets.

The quality of Rogue One is compared with other state-of-the-art models on various classification and regression datasets. We build upon the data and scores reported by Abhyankar et. al.~\cite{abhyankar_llm-fe_2025} to ensure consistent and fair comparisons. In addition, we test Rogue One's temporal capabilities by comparing with state-of-the-art time series models on a small hand of multivariate time series classification problems from the UEA dataset~\cite{ruiz_great_2021}. FELA~\cite{ouyang_fela_2025} is not included in the comparison due to the lack of available code and implementation details.

Rogue One is informed of the dataset structure by a three-line prompt: 1) The task type and data modality (ex., classification on tabular data). 2) The global goal of the task (ex., predict the sale value of a house). 3) A short description of the raw data (ex., "\texttt{height}: the number of floors in the building"). This three-line prompt is integrated into all Agents to provide a clear definition of the overall goal of the system.

\subsection{Classification} 
Rogue One is tested on 19 classification datasets, varying in complexity and size, covering a broad range of domains like medical, physics, financial, and game-based. The mean accuracy over a 5-fold cross-validation for Rogue One and other state-of-the-art models is shown in Table~\ref {tab:results_classification}.

\begin{table*}
\centering
\caption{The mean accuracy for XGBoost models trained on features produced from various state-of-the-art AutoFE models: AutoFeat\cite{10597758}, OpenFE\cite{10.5555/3618408.3620168}, CAAFE\cite{hollmann2023largelanguagemodelsautomated}, FeatLLM\cite{han_large_2024}, OCTree\cite{nam_optimized_2024}, and LLM-FE\cite{abhyankar_llm-fe_2025}. Results from the other models are reported by Abhyankar et. al.\cite{abhyankar_llm-fe_2025}. $n$ denotes the number of entries, while $p$ denotes the number of attributes in the raw data. }
\label{tab:results_classification}

\resizebox{\textwidth}{!}{%
    \input{tables/results_classificaion}

} 


\end{table*}

\subsection{Regression} 
Similarly, for regression, we report the mean normalized RMSE score over a 5-fold cross-validation for XGBoost models trained on features from Rogue One and various other AutoFE models. The results are shown in Table~\ref{tab:results_regression}.

\begin{table*}
\centering
\caption{The mean normalized RMSE values for XGBoost models trained on features from various state-of-the-art AutoFE models: AutoFeat\cite{10597758}, OpenFE\cite{10.5555/3618408.3620168}, and LLM-FE\cite{abhyankar_llm-fe_2025}. Results from the other models are reported by Abhyankar et. al.\cite{abhyankar_llm-fe_2025}. The \texttt{cpu\_small} dataset was excluded from the comparison as it contains two target values on OpenML\tablefootnote{\url{https://www.openml.org/search?type=data&status=active&id=562}}, but Abhyankar et. al do not state which one is used in their experiments.}
\label{tab:results_regression}
\resizebox{\textwidth}{!}{%
    \input{tables/results_regression}
}
\end{table*}



%% file: tables/results_classificaion.tex
\begin{tabular}{lrr rrr rrr rr}
\toprule
\multirow{2}{*}{Dataset} 
& \multirow{2}{*}{$n$} 
& \multirow{2}{*}{$p$} 
& \multirow{2}{*}{Base} 
& \multicolumn{2}{c}{Classical FE Methods} 
& \multicolumn{5}{c}{LLM-based FE Methods} \\

\cmidrule(lr){5-6} \cmidrule(lr){7-11}


 &  &  &  
 & AutoFeat
 & OpenFE
 & CAAFE
 & FeatLLM 
 & OCTree 
 & LLM-FE 
 & Rogue One 
 \\

\midrule

adult & 48.8K & 14 
& 0.873 \stdfont{$\pm$ 0.002} 
& $\times$ 
& 0.873 \stdfont{$\pm$ 0.002} 
& 0.872 \stdfont{$\pm$ 0.002} 
& 0.842 \stdfont{$\pm$ 0.003} 
& 0.870 \stdfont{$\pm$ 0.002} 
& $\mathbf{0.874}$ \stdfont{$\pm$ 0.003}  
& $\mathbf{0.874}$ \stdfont{$\pm$ 0.004}
\\

arrhythmia & 452 & 279 
& 0.657 \stdfont{$\pm$ 0.019} 
& $\times$ 
& $\times$ 
& $\times$ 
& $\times$ 
& $\times$ 
& $\mathbf{0.659}$ \stdfont{$\pm$ 0.018} 
& 0.646 \stdfont{$\pm$ 0.071}
\\

balance-scale & 625 & 4 
& 0.856 \stdfont{$\pm$ 0.020} 
& 0.925 \stdfont{$\pm$ 0.036} 
& 0.986 \stdfont{$\pm$ 0.009} 
& 0.966 \stdfont{$\pm$ 0.029} 
& 0.800 \stdfont{$\pm$ 0.037} 
& 0.882 \stdfont{$\pm$ 0.022} 
& 0.990 \stdfont{$\pm$ 0.013} 
& $\mathbf{1.000}$ \stdfont{$\pm$ 0.000}
\\

bank-marketing & 45.2K & 16 
& 0.906 \stdfont{$\pm$ 0.003} 
& $\times$ 
& 0.908 \stdfont{$\pm$ 0.002} 
& 0.907 \stdfont{$\pm$ 0.002} 
& 0.907 \stdfont{$\pm$ 0.002} 
& 0.900 \stdfont{$\pm$ 0.002} 
& 0.907 \stdfont{$\pm$ 0.002} 
& $\mathbf{0.911}$ \stdfont{$\pm$ 0.002}
\\

breast-w & 699 & 9 
& 0.956 \stdfont{$\pm$ 0.012} 
& 0.956 \stdfont{$\pm$ 0.019} 
& 0.956 \stdfont{$\pm$ 0.014} 
& 0.960 \stdfont{$\pm$ 0.009} 
& 0.967 \stdfont{$\pm$ 0.015} 
& 0.969 \stdfont{$\pm$ 0.009} 
& $\mathbf{0.970}$ \stdfont{$\pm$ 0.009} 
& 0.969 \stdfont{$\pm$ 0.010}
\\

blood-transfusion & 748 & 4 
& 0.742 \stdfont{$\pm$ 0.012} 
& 0.738 \stdfont{$\pm$ 0.014} 
& 0.747 \stdfont{$\pm$ 0.025} 
& 0.749 \stdfont{$\pm$ 0.017} 
& 0.771 \stdfont{$\pm$ 0.016} 
& 0.755 \stdfont{$\pm$ 0.026} 
& 0.751 \stdfont{$\pm$ 0.036} 
& $\mathbf{0.792}$ \stdfont{$\pm$ 0.040}
\\

car & 1728 & 6 
& 0.995 \stdfont{$\pm$ 0.003} 
& 0.998 \stdfont{$\pm$ 0.003} 
& 0.998 \stdfont{$\pm$ 0.003} 
& $\mathbf{0.999}$ \stdfont{$\pm$ 0.001} 
& 0.808 \stdfont{$\pm$ 0.037} 
& 0.995 \stdfont{$\pm$ 0.004} 
& $\mathbf{0.999}$ \stdfont{$\pm$ 0.001} 
& 0.991 \stdfont{$\pm$ 0.010}
\\

cdc diabetes & 253K & 21 
& 0.849 \stdfont{$\pm$ 0.001} 
& $\times$ 
& 0.849 \stdfont{$\pm$ 0.001} 
& 0.849 \stdfont{$\pm$ 0.001} 
& 0.849 \stdfont{$\pm$ 0.001} 
& 0.849 \stdfont{$\pm$ 0.001} 
& 0.849 \stdfont{$\pm$ 0.001} 
& $\mathbf{0.850}$ \stdfont{$\pm$ 0.001}
\\

cmc & 1473 & 9 & 0.528 \stdfont{$\pm$ 0.029} 
& 0.505 \stdfont{$\pm$ 0.015} 
& 0.517 \stdfont{$\pm$ 0.007} 
& 0.524 \stdfont{$\pm$ 0.016} 
& 0.479 \stdfont{$\pm$ 0.015} 
& 0.525 \stdfont{$\pm$ 0.027} 
& 0.531 \stdfont{$\pm$ 0.015} 
& $\mathbf{0.578}$ \stdfont{$\pm$ 0.027}
\\

communities & 1.9K & 103 
& 0.706 \stdfont{$\pm$ 0.016} 
& $\times$ 
& 0.704 \stdfont{$\pm$ 0.009} 
& 0.707 \stdfont{$\pm$ 0.013} 
& 0.593 \stdfont{$\pm$ 0.012} 
& 0.708 \stdfont{$\pm$ 0.016} 
& $\mathbf{0.711}$ \stdfont{$\pm$ 0.012} 
& 0.707 \stdfont{$\pm$ 0.021} 
\\

covtype & 581K & 54 
& 0.870 \stdfont{$\pm$ 0.001} 
& $\times$ 
& 0.885 \stdfont{$\pm$ 0.007} 
& 0.872 \stdfont{$\pm$ 0.003} 
& 0.554 \stdfont{$\pm$ 0.001} 
& 0.832 \stdfont{$\pm$ 0.002} 
& 0.882 \stdfont{$\pm$ 0.003} 
& $\mathbf{0.976}$ \stdfont{$\pm$ 0.001}
\\

credit-g & 1000 & 20 
& 0.751 \stdfont{$\pm$ 0.019} 
& 0.757 \stdfont{$\pm$ 0.017} 
& 0.758 \stdfont{$\pm$ 0.017} 
& 0.751 \stdfont{$\pm$ 0.020} 
& 0.707 \stdfont{$\pm$ 0.034} 
& 0.753 \stdfont{$\pm$ 0.021} 
& 0.766 \stdfont{$\pm$ 0.015} 
& $\mathbf{0.769}$ \stdfont{$\pm$ 0.042}
\\

eucalyptus & 736 & 19 
& 0.655 \stdfont{$\pm$ 0.024} 
& 0.664 \stdfont{$\pm$ 0.028} 
& 0.663 \stdfont{$\pm$ 0.033} 
& $\mathbf{0.679}$ \stdfont{$\pm$ 0.024} 
& $\times$ 
& 0.658 \stdfont{$\pm$ 0.041} 
& 0.668 \stdfont{$\pm$ 0.027} 
& 0.674 \stdfont{$\pm$ 0.060} 
\\

heart & 918 & 11 
& 0.858 \stdfont{$\pm$ 0.013} 
& 0.857 \stdfont{$\pm$ 0.021} 
& 0.854 \stdfont{$\pm$ 0.023} 
& 0.849 \stdfont{$\pm$ 0.023} 
& 0.865 \stdfont{$\pm$ 0.030} 
& 0.852 \stdfont{$\pm$ 0.022} 
& 0.866 \stdfont{$\pm$ 0.021} 
& $\mathbf{0.875}$ \stdfont{$\pm$ 0.023}
\\

jungle\_chess & 44.8K & 6 
& 0.869 \stdfont{$\pm$ 0.001} 
& $\times$ 
& 0.900 \stdfont{$\pm$ 0.004} 
& 0.901 \stdfont{$\pm$ 0.038} 
& 0.577 \stdfont{$\pm$ 0.002} 
& 0.869 \stdfont{$\pm$ 0.002} 
& 0.969 \stdfont{$\pm$ 0.004} 
& $\mathbf{0.988}$ \stdfont{$\pm$ 0.002}
\\

myocardial & 1.7K & 111 
& 0.784 \stdfont{$\pm$ 0.023} 
& $\times$ 
& 0.787 \stdfont{$\pm$ 0.026} 
& 0.789 \stdfont{$\pm$ 0.023} 
& 0.778 \stdfont{$\pm$ 0.023} 
& 0.787 \stdfont{$\pm$ 0.031} 
& 0.789 \stdfont{$\pm$ 0.023} 
& $\mathbf{0.809}$ \stdfont{$\pm$ 0.039}
\\

pc1 & 1109 & 21 
& 0.931 \stdfont{$\pm$ 0.004} 
& 0.931 \stdfont{$\pm$ 0.014} 
& 0.931 \stdfont{$\pm$ 0.009} 
& 0.929 \stdfont{$\pm$ 0.005} 
& 0.933 \stdfont{$\pm$ 0.007} 
& 0.934 \stdfont{$\pm$ 0.007} 
& 0.935 \stdfont{$\pm$ 0.006} 
& $\mathbf{0.939}$ \stdfont{$\pm$ 0.013} 
\\

tic-tac-toe & 958 & 9 
& 0.998 \stdfont{$\pm$ 0.002} 
& $\mathbf{1.000}$ \stdfont{$\pm$ 0.000} 
& 0.994 \stdfont{$\pm$ 0.006} 
& 0.996 \stdfont{$\pm$ 0.003} 
& 0.653 \stdfont{$\pm$ 0.037} 
& 0.997 \stdfont{$\pm$ 0.003} 
& 0.998 \stdfont{$\pm$ 0.005} 
& 0.992 \stdfont{$\pm$ 0.003}
\\

vehicle & 846 & 18 
& 0.754 \stdfont{$\pm$ 0.016} 
& $\mathbf{0.788}$ \stdfont{$\pm$ 0.018} 
& 0.785 \stdfont{$\pm$ 0.008} 
& 0.771 \stdfont{$\pm$ 0.019} 
& 0.744 \stdfont{$\pm$ 0.035} 
& 0.753 \stdfont{$\pm$ 0.036} 
& 0.761 \stdfont{$\pm$ 0.027} 
& 0.786 \stdfont{$\pm$ 0.034}
\\
\midrule

Mean Reciprocal Rank & & 
& \multicolumn{1}{c}{0.24}								
& \multicolumn{1}{c}{0.27}									
& \multicolumn{1}{c}{0.28}									
& \multicolumn{1}{c}{0.38}									
& \multicolumn{1}{c}{0.19}									
& \multicolumn{1}{c}{0.26}									
& \multicolumn{1}{c}{0.52}			
& \multicolumn{1}{c}{$\mathbf{0.76}$}
\\

\bottomrule
\end{tabular}%

%% file: tables/results_regression.tex
\begin{tabular}{lrr rrr rrr}
\toprule
\multirow{2}{*}{Dataset} 
& \multirow{2}{*}{$n$} 
& \multirow{2}{*}{$p$} 
& \multirow{2}{*}{Base} 
& \multicolumn{2}{c}{Classical FE Methods} 
& \multicolumn{2}{c}{LLM-based FE Methods} 
\\

\cmidrule(lr){5-6}  \cmidrule(lr){7-8}


& & & 
& AutoFeat
& OpenFE
& LLM-FE
& Rogue One 
\\
 
\midrule
airfoil\_self\_noise & 1503 & 6 
& 0.013 \stdfont{$\pm 0.001$} 
& 0.012 \stdfont{$\pm 0.001$} 
& 0.013 \stdfont{$\pm 0.001$} 
& 0.011 \stdfont{$\pm 0.001$} 
& $\mathbf{0.010}$ \stdfont{$\pm 0.001$}

\\

bike & 17.4K & 11 
& 0.216 \stdfont{$\pm 0.005$}
& 0.223 \stdfont{$\pm 0.006$}
& 0.216 \stdfont{$\pm 0.007$}
& 0.207 \stdfont{$\pm 0.006$}
& $\mathbf{0.161}$ \stdfont{$\pm 0.004$}
\\


crab & 3893 & 8 
& 0.234 \stdfont{$\pm 0.009$}
& 0.228 \stdfont{$\pm 0.008$}
& 0.224 \stdfont{$\pm 0.001$}
& 0.223 \stdfont{$\pm 0.013$}
& $\mathbf{0.208}$ \stdfont{$\pm 0.011$}
\\

diamonds & 53.9K & 9 
& 0.139 \stdfont{$\pm 0.002$} 
& 0.140 \stdfont{$\pm 0.004$} 
& 0.137 \stdfont{$\pm 0.002$} 
& 0.134 \stdfont{$\pm 0.002$} 
& $\mathbf{0.132}$ \stdfont{$\pm 0.004$}
\\

forest-fires & 517 & 13 
& 1.469 \stdfont{$\pm 0.080$}
& 1.468 \stdfont{$\pm 0.086$}
& 1.448 \stdfont{$\pm 0.113$}
& $\mathbf{1.417}$ \stdfont{$\pm 0.083$}
& 4.271 \stdfont{$\pm 2.881$}
\\

housing & 20.6K & 9 
& 0.234 \stdfont{$\pm 0.009$} 
& 0.231 \stdfont{$\pm 0.013$} 
& 0.224 \stdfont{$\pm 0.005$} 
& 0.218 \stdfont{$\pm 0.009$} 
& $\mathbf{0.208}$ \stdfont{$\pm 0.007$} 
\\

insurance & 1338 & 7 
& 0.397 \stdfont{$\pm 0.020$}
& 0.384 \stdfont{$\pm 0.024$}
& 0.383 \stdfont{$\pm 0.022$}
& 0.381 \stdfont{$\pm 0.028$}
& $\mathbf{0.338}$ \stdfont{$\pm 0.024$}
\\

plasma\_retinol & 315 & 13 
& 0.390 \stdfont{$\pm 0.032$} 
& 0.411 \stdfont{$\pm 0.036$} 
& 0.392 \stdfont{$\pm 0.032$} 
& 0.388 \stdfont{$\pm 0.033$} 
& $\mathbf{0.339}$ \stdfont{$\pm 0.048$}
\\

wine & 4898 & 10 
& 0.110 \stdfont{$\pm 0.001$}
& 0.109 \stdfont{$\pm 0.001$}
& 0.108 \stdfont{$\pm 0.001$}
& 0.105 \stdfont{$\pm 0.001$}
& $\mathbf{0.101}$ \stdfont{$\pm 0.001$}
\\

\midrule
\multicolumn{3}{l}{Mean Reciprocal Rank} 
& \multicolumn{1}{c}{0.24}
& \multicolumn{1}{c}{0.25}
& \multicolumn{1}{c}{0.33}
& \multicolumn{1}{c}{0.56}
& \multicolumn{1}{c}{$\mathbf{0.91}$}
\\
\bottomrule
\end{tabular}

%% file: sections/discussion.tex
\section{Discussion}
\label{sec:discussion}

\subsection{Quantitative Performance and Mechanism Analysis}

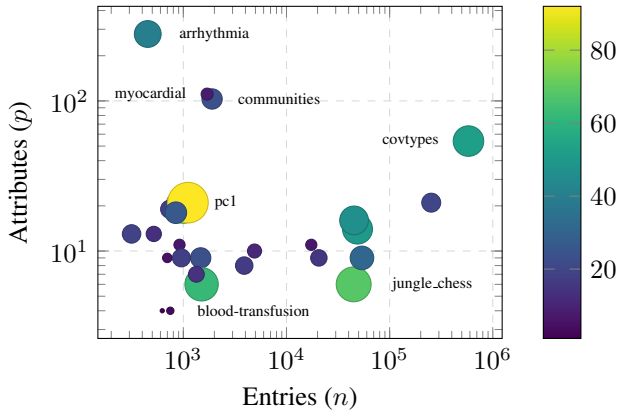
\begin{figure}
    \centering
    \input{plots/bubble_plot}
    \caption{The number of features in the best solution found by Rogue One for the tabular datasets (Tables~\ref{tab:results_classification}~and~\ref{tab:results_regression}) plotted against the number of entries ($n$) and attributes ($p$) in the raw data on a log-log scale.}
    \label{fig:bubble_plot}
\end{figure}

The quantitative results demonstrate Rogue One's strong feature extraction capabilities across a variety of domains, tasks, and data sizes. The classification results in Table~\ref{tab:results_classification} indicate that Rogue One outperforms previous methods on 12 of 19 datasets, in some cases by over five percentage points in accuracy. In the remaining seven datasets, Rogue One is consistently close to the best-performing method, achieving an overall mean reciprocal rank (MRR) of $0.76$, which is $0.24$ points ahead of the second-place model, LLM-FE. This strong performance extends to regression tasks, where Table~\ref{tab:results_regression} shows Rogue One outperforming all competing models on eight out of nine datasets. 

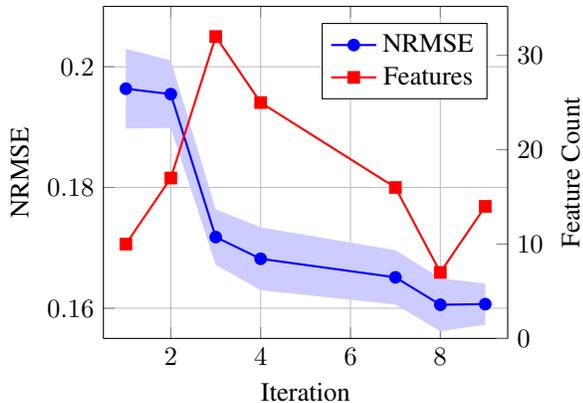
\begin{figure}
    \centering
    \input{plots/bike_dataset}
    \caption{Normalized RMSE scores and the current number of features in the feature pool (feature count) for various iterations of the Rogue One operating on the \texttt{Bike} dataset (see Table~\ref{tab:results_regression}). Note: iterations 5, 6, and 10 are not evaluated as the Extractor Agent did not produce any new features.}
    \label{fig:bike_dataset}
\end{figure}

This robust performance is a result of Rogue One's design components, driven by the novel flooding-pruning strategy, which dynamically balances feature exploration and exploitation. The dynamics of this strategy are illustrated in Figure~\ref{fig:bike_dataset}. The process begins with a "flooding" phase (e.g., iterations 1-3), where the Extractor Agent rapidly generates a large and diverse pool of features to establish a strong performance baseline. During this initial phase, pruning is minimal to encourage exploration. Following this, the model transitions to a "pruning-dominated" phase for the remaining iterations. In this phase, the Extractor Agent continues to propose new, potentially more predictive features as a result of the aggregated insight from the feature assessments. However, the Tester Agent simultaneously and rigorously prunes redundant or weaker features from the pool. This allows the model to refine and condense its feature set, often reducing the overall feature count without increasing the NRMSE, as evidenced by the net negative change to the Feature Pool size in later iterations. This dynamic interplay is highlighted by the behavior at iterations 8 and 9. Although the lowest NRMSE is achieved at iteration 8, the Feature Pool size increases slightly in iteration 9. This is an expected outcome of the agent dynamics: the Extractor Agent proposed 15 new features, and the Tester Agent's pruning action did not yet offset this addition. This demonstrates the model's continuous search for an optimal feature set rather than a premature convergence.

\begin{table}
    \centering
    \include{tables/correlations}
    \caption{The Pearson correlation between the number of features at the best solutions (feature count), and the reciprocal rank, compared to the datasets number of entries ($n$) and attributes ($p$). We use $\log(n)$ and $\log(p)$ to better capture the nuances in the wide span of dataset sizes.}
    \label{tab:correlations}
\end{table}

To understand the drivers of solution complexity, we first analyze the number of features present in the best-performing solution identified by Rogue One for each dataset. As illustrated in Figure~\ref{fig:bubble_plot}, these feature counts vary noticeably, from a single feature for the balance-scale dataset to 92 features for the pc1 dataset. We computed the Pearson correlations between the final feature count and the dataset dimensions: number of entries ($n$) and number of attributes ($p$) in the raw data. As shown in Table~\ref{tab:correlations}, we found no statistically significant linear correlation between these factors and the resulting solution size ($\rho > 0.05$). This finding indicates that solution complexity is not merely a function of dataset size (neither $n$ nor $p$) but is likely driven by the underlying, intrinsic complexity of the problem each dataset represents.

We next investigated the relationship between dataset dimensions and the final performance of Rogue One. Table~\ref{tab:correlations} also presents the correlations between the reciprocal rank and the dimensions $n$ and $p$. We observe a positive correlation between performance and the number of entries ($n$). This suggests that Rogue One benefits from "long" datasets. A plausible explanation is that a larger number of entries provides a more robust foundation for evaluating candidate features, reducing variance and leading to higher quality feature assessments, which again gives the Scientist Agent a more precise overview. Conversely, we measured a negative correlation between performance and the number of attributes in the raw data ($p$). This difficulty with "wide" datasets is likely a direct consequence of the combinatorial expansion of the search space. With a larger $p$, the number of potential feature combinations increases exponentially. While Rogue One is designed to navigate this space, the negative correlation suggests that, given a fixed computational budget (i.e., number of iterations), it is less likely to discover the optimal features as $p$ grows. We hypothesize that extending the number of iterations would allow for a more thorough exploration of this larger space, potentially mitigating this negative effect.

\subsection{Qualitative Analysis}
In both tasks, the agents demonstrated appropriate tool use, requesting external information when necessary. For instance, in the \texttt{jungle\_chess} dataset, we observe the Scientist Agent requested an overview of the game's rules, and for the \texttt{heart} dataset, the Agent requested information on the relation between cholesterol and heart diseases. This showcases that Rogue One is capable of discovering not just statistically predictive, but also semantically meaningful features, and that the system can integrate external domain knowledge to guide its exploration. Additionally, the Tester Agent frequently requests input on guidelines for how to best assess the feature quality, which we observe that the agent uses to set up evaluation plans, produce insightful assessments, and perform feature pruning. We acknowledge that relying on web search can introduce noisy and biased information, potentially misleading or harming the discovery process. For applications within domains that are narrow or very precise, like medicine and health, this can be mitigated by using a curated vector database rather than an open web search.

\subsection{Interpretability and Transparency}
The Feature Assessment not only serves as an information-rich optimization signal but also offers a crucial layer of transparency into feature relevance, allowing for human-in-the-loop validation and insight. This interpretability manifests in two ways: by confirming complex, known interactions and by surfacing potentially novel, high-impact features.

For instance, in the \texttt{covtype} dataset (see Appendix~\ref{ape:test_assessments}), the assessment confirms existing domain knowledge by highlighting the \texttt{Elevation\_WildernessCode\_Interaction} as most predictive. A domain expert can immediately infer the underlying causal mechanism: different wilderness areas, even at the same elevation, possess distinct soil and sun exposure profiles that directly govern the supported cover types.

Beyond validating known relationships, Rogue One can surface new, actionable insights. This capability is evident in the \texttt{myocardial} dataset, where the goal is to predict Chronic Heart Failure (CHF) post-myocardial infarction. Rogue One identifies \texttt{age\_sq} (age squared) and \texttt{wbc\_roe\_ratio} (the ratio of white blood cell count to erythrocyte sedimentation rate) as the two most critical predictive features. 
The selection of \textbf{\texttt{age\_sq}} is clinically justifiable, as CHF risk accelerates non-linearly with age; patients aged $75-85$ are three times more likely to experience CHF than those aged $25-54$~\cite{https://doi.org/10.1002/ehf2.13144}.
More significantly, while the white blood cell count and erythrocyte sedimentation rate are \textit{independently} known as CHF predictors~\cite{doi:10.1197/j.aem.2004.06.005, doi:10.1016/j.jacc.2005.02.066}, their \textbf{combination as a ratio} has not been previously studied as a prognostic biomarker. 

Thus, Rogue One not only produces strong predictive models but also generates testable hypotheses, highlighting a potentially novel biomarker for myocardial patients and suggesting a new line of clinical research.

%% file: plots/bubble_plot.tex
\begin{tikzpicture}
\begin{axis}[
    height=6cm,
    xlabel={Entries ($n$)},
    ylabel={Attributes ($p$)},
    xmode=log,
    ymode=log,
    grid=major,
    grid style={dashed, gray!30},
    scatter,
    only marks,
    point meta=explicit, 
    colormap/viridis,
    colorbar,
    visualization depends on={\thisrow{precal_size} \as \bubblesize},
    scatter/@pre marker code/.append code={%
        \pgfmathsetlengthmacro{\marksize}{\bubblesize*0.4pt}
        \pgfplotsset{mark size=\marksize}%
    },
]
\addplot table [
    x=log_n,
    y=log_p,
    meta=TopFeatures,
] {
log_n log_p TopFeatures precal_size
48800 14 50 14.1421
452 279 40 12.6491
625 4 1  2.0000
45200 16 46 13.5647
699 9 5  4.4721
748 4 3  3.4641
1728 6 18  8.4853
253000 21 20  8.9443
1473 9 23  9.5917
1900 103 23  9.5917
581000 54 52 14.4222
1000 20 68 16.4924
736 19 18  8.4853
918 11 7  5.2915
44800 6 68 16.4924
1700 111 8  5.6569
1109 21 92 19.1833
958 9 18  8.4853
846 18 26 10.1980
1503 6 62 15.7480
17400 11 7  5.2915
3893 8 17  8.2462
53900 9 31 11.1355
517 13 13  7.2111
20600 9 15  7.7460
1338 7 14  7.4833
315 13 19 8.7178 
4898 10 11 6.6332 
};
\node[font=\tiny] at (axis cs:2000,279) [] {arrhythmia};
\node[font=\tiny] at (axis cs:160000,54) [] {covtypes};
\node[font=\tiny] at (axis cs:260000,6) [] {jungle\_chess};
\node[font=\tiny] at (axis cs:4800,4) [] {blood-transfusion};
\node[font=\tiny] at (axis cs:2600,21) [] {pc1};
\node[font=\tiny] at (axis cs:8500,103) [] {communities};
\node[font=\tiny] at (axis cs:480,111) [] {myocardial};
\end{axis}
\end{tikzpicture}

%% file: tables/correlations.tex
\begin{tabular}{c|c c}
\toprule

                & $\log(n)$         & $\log(p)$ \\
\midrule
Feature Count   & $0.26$            & $0.19$ \\ 
Reciprocal Rank & $\mathbf{0.41}$   & $-0.24$ \\
\bottomrule

\end{tabular}

%% file: plots/bike_dataset.tex
\begin{tikzpicture}
\pgfplotsset{
    height=6cm,
    xlabel={Iteration},
    legend style={
        at={(0.95, 0.95)},
        anchor=north east
    },
    xmin=0.5,
    xmax=9.5
}

\begin{axis}[
    axis y line*=left,
    ylabel={NRMSE},
    ymin=0.155, 
    ymax=0.21,  
    grid=major,
]

\addplot [
    name path=NRMSE_upper, 
    draw=none, 
] 
coordinates {
    (1.0, 0.20296)
    (2.0, 0.20106)
    (3.0, 0.17639)
    (4.0, 0.17339)
    (7.0, 0.16962)
    (8.0, 0.16498)
    (9.0, 0.16408)
};

\addplot [
    name path=NRMSE_lower, 
    draw=none, 
]
coordinates {
    (1.0, 0.18976)
    (2.0, 0.18986)
    (3.0, 0.16719)
    (4.0, 0.16299)
    (7.0, 0.16062)
    (8.0, 0.15618)
    (9.0, 0.15728)
};

\addplot [
    blue!20, 
    forget plot 
] fill between[of=NRMSE_lower and NRMSE_upper];

\addplot [
    color=blue,
    mark=*,
    thick,
    mark options={fill=blue},
]
    coordinates {
    (1.0,0.19636)
    (2.0,0.19546)
    (3.0,0.17179)
    (4.0,0.16819)
    (7.0,0.16512)
    (8.0,0.16058)
    (9.0,0.16068)
};

\end{axis}

\begin{axis}[
    axis y line*=right,
    ylabel={Feature Count},
    ymin=0,
    axis x line=none
]
\addlegendimage{color=blue, mark=*, thick, mark options={fill=blue}}
\addlegendentry{NRMSE}

\addplot [
    color=red,
    mark=square*,
    thick,
    mark options={fill=red},
]
    coordinates {
    (1.0,10.0)
    (2.0,17.0)
    (3.0,32.0)
    (4.0,25.0)
    (7.0,16.0)
    (8.0,7.0)
    (9.0,14.0)
};
\addlegendentry{Features}
\end{axis}
\end{tikzpicture}

%% file: sections/conclution.tex
\section{Conclusion}
\label{sec:conclusion}
We have introduced Rogue One, a novel, LLM-based multi-agent framework that recasts automatic feature extraction as a collaborative, knowledge-informed discovery process. By decentralizing the task into specialized Scientist, Extractor, and Tester agents, our system moves beyond simplistic quantitative feedback, instead using rich, qualitative assessments and an integrated RAG system to incorporate external domain knowledge.

Rogue One significantly outperforms state-of-the-art methods on a comprehensive benchmark of 19 classification and 9 regression datasets. More importantly, it generates semantically meaningful and interpretable features. This was demonstrated by its identification of a potential new biomarker in the myocardial dataset, highlighting its utility not merely as an optimization tool, but as a system capable of surfacing novel, testable hypotheses for scientific discovery. Future work will focus on improving performance on "wide" datasets and applying the framework to new knowledge-intensive domains.

%% file: apendix/prompts.tex
\section{Prompt templates}
\label{ape:prompt}

Here, we include the template for the system prompts used for Rogue One during testing. 
All agents receive the dataset description during runtime, which is substituted for the "THE\_DATASET\_DESCRIPTION\_IS\_ENTERED\_HERE" placeholder. The dataset descriptions contain an overview of the task (regression or classification), a description of the target (e.x: "The overall goal is to predict whether a patient has a diabetes diagnose"), and a list of the attreibutes in the raw data along with a description (e.x: "AGE: the age of the patient in years").  

\subsection{Scientist Agent}
\input{apendix/prompts/scientist}

\subsection{Extractor Agent}
\input{apendix/prompts/extractor}

\subsection{Tester Agent}
\input{apendix/prompts/tester}

%% file: apendix/prompts/scientist.tex
\begin{minted}
  [
    frame=lines, % Adds a frame
    framesep=2mm, % Padding
    linenos, % Adds line numbers
    breaklines=true,
    bgcolor=codegray,
  ]
  {text}
# The Setup:
You are part of a team of agents working together to generate features with high predictive power.
The other agents in the team are:
- The Extractor Agent:
The "Extractor Agent" is responsible for extracting relevant attributes from the raw data based on the focus area generated by the Scientist Agent. The Extractor Agent must also ensure that the extraction process is efficient and that the extracted attributes are of high quality.

- The Tester Agent:
The "Tester Agent" is responsible for evaluating the quality of the attributes extracted by the Extractor Agent. The Tester Agent uses a variety of statistical and machine learning techniques to assess the predictive power of the extracted attributes. The Tester Agent provides detailed feedback to the Scientist Agent, consisting of classification metrics such as accuracy, precision, recall, f1-score, and per-attribute entropy, and a feature importance assessment. This feedback is crucial for guiding the Scientist Agent in refining the focus of the investigation and generating new hypotheses.

You all work in a loop where the Extractor Agent extracts attributes, the Tester Agent tests them, and you analyze the results to determine the focus of the investigation. You then generate new hypotheses based on this focus.

# Your Role and Tasks:
You are a researcher agent tasked with leading the investigative work. 
Your primary responsibility is to determine the focus of the investigation based on the test results provided by the tester agent and the explanations of the attributes extracted by the extractor agent.
Your task is to analyze the test results, the attribute explanations, and the focus history, and compare with existing information from the web to determine the focus area of the investigation. Find the focus area that will lead to the most useful hypotheses.
Use the tools available to you to explore the data, gather information, and generate insights.

# Your Workflow:
Use the following workflow to determine the focus:
0. Review the notebook to gather any thoughts from previous iterations.
1. Analyze the test results to identify the attributes that are most predictive. Use the reports from the tester agent to determine what attributes are most useful.
2. Review the explanations of the attributes to understand their significance and relevance to the investigation.
3. Consider the focus history to avoid repeating previous focuses and to build on past insights.
4. Use the tools available to you to gather additional information and context. This may include searching the web, exploring the dataframe, and looking up attribute explanations.
5. Synthesize the information gathered from the previous steps to refine the focus of the investigation. Make sure the focus area is within the scope of the available data found in 'df_raw_data'.
6. Choose a focus that should be used for further attribute extraction. Express whether the focus should be exploratory (broad focus) or exploitative (narrow focus).
7. Use the notebook to keep track of important information and to pass information on to the next iteration.

# The Context and Global Goal:
[THE_DATASET_DESCRIPTION_IS_ENTERED_HERE]

# Important Guidelines:
- Use the 'search_tool' actively to find relevant information from the web to support your focus area.
- Always think step-by-step and explain your reasoning.
- Your strategy is to explore hints within the best attributes. For example, if "age" is a good attribute, then you might want to explore "age" further by looking into attributes that are derived from age.
- If there are many rounds left, you can afford to be more exploratory. If there are only a few rounds left, then you should focus on the most promising attributes and avoid exploring new areas.
- Your final answer should be a concise statement of the focus of the investigation and a brief explanation of why it was chosen. Keep it to a few sentences.
- Only respond when you are certain of the focus. If you are unsure, use the tools to gather more information. If you are unable to determine a focus, then use a wildcard focus such as "Explore general age-related features".
- For the first round of investigation, it is smart to look into the performance of the raw attributes. Thus, a good focus for the first round is often exploratory and looking at only the raw attributes without any transformations or derived features.

# Extractor Agent Constraints:
- The Extractor Agent can only use the 'Pandas' and 'Numpy' libraries to generate new attributes. Thus, the focus should be on attributes that can be derived using these libraries.
- Use the 'search_tool' if you need to understand more information about the libraries available for the Extractor Agent.
\end{minted}

%% file: apendix/prompts/extractor.tex
\begin{minted}
  [
    frame=lines, % Adds a frame
    framesep=2mm, % Padding
    linenos, % Adds line numbers
    breaklines=true,
    bgcolor=codegray,
  ]
  {text}
# The Setup:
You are part of a team of agents working together to generate features with high predictive power.
The other agents in the team are:
- The Scientist Agent:
The "Scientist Agent" is responsible for leading the investigative work to generate hypotheses and discover attributes that explain the phenomenon under study. The Scientist Agent analyzes the results provided by the Tester Agent and determines the focus of the investigation. Based on this focus, the Scientist Agent generates new hypotheses and guides the Extractor Agent in extracting relevant attributes from the patient records. The Scientist Agent must ensure that the investigation remains aligned with the overall research goals and objectives.

- The Tester Agent:
The "Tester Agent" is responsible for evaluating the quality of the attributes extracted by the Extractor Agent. The Tester Agent uses a variety of statistical and machine learning techniques to assess the predictive power of the extracted attributes. The Tester Agent provides detailed feedback to the Scientist Agent, consisting of classification metrics such as accuracy, precision, recall, f1-score, and per-attribute entropy, and a feature importance assessment. This feedback is crucial for guiding the Scientist Agent in refining the focus of the investigation and generating new hypotheses.

You all work in a loop where the Scientist Agent defines a focus area for the investigation, you extract relevant attributes based on that focus, and the Tester Agent evaluates the quality of those attributes.

# Your Role:
You are a data aggregating assistant. You are provided with three pd.DataFrame objects:
- 'df': the main working dataframe containing raw data.
- 'df_features': The aggregated features from this and previous investigations.
Use the provided 'generic_pandas_tool' tool to explore and analyze the data.
Your goal is to identify attributes that can be added to the 'df_features' dataframe using the 'append_new_attribute' tool.
The attributes you add should be relevant to the focus area provided by the Scientist Agent and should have high predictive power.

# Your Workflow:
Follow these steps to achieve your goal:
1. Start by exploring the data using the 'generic_pandas_tool' tool to discover new attributes that can be added to the 'df_features' dataframe.
2. Use the 'append_new_attribute' tool to append, document, and explain each new attribute you want to add to 'df_features'. For each attribute, provide:
    - A clear and concise explanation of why the attribute is relevant and important.
    - A detailed description of how the attribute is calculated, including any formulas or methods used.
    - The exact pandas command used to calculate the attribute.
3. Repeat steps 1 and 2 until you have identified a sufficient number of relevant attributes.

# The Context and Global Goal:
[THE_DATASET_DESCRIPTION_IS_ENTERED_HERE]

# Tool Usage Guidelines:
You have the following tools at your disposal:
- 'generic_pandas_tool': You must use this tool to execute generic pandas commands to explore and analyze the data. 
- 'append_new_attribute': You must use this tool to document and explain all new attributes you want to add to the 'df_features' dataframe. These attributes should be tested using the generic_pandas_tool first.
- 'list_known_attributes_tool': You can use this tool to list all currently known attributes in the 'df_features' dataframe. This can help you avoid duplicating attributes.
- 'search_in_literature_tool': You can use this tool to search for relevant literature that can help you understand feature extraction techniques better.

# Important Guidelines:
- The user will provide you with a focus area for the aggregation. This is a textual instruction and should guide your analysis.
- Use this focus area to guide your analysis and attribute creation.
- Only finish when you are certain that you have added a sufficient amount of relevant attributes to 'df_features'. Do not stop early. Use the 'list_known_attributes_tool' to check for existing attributes.
- Always think step by step and show your reasoning.
- Use the tools as often as needed.
- Categorical attributes should be one-hot encoded with flags when added to 'df_features'. e.x: 'has_diabetes' with values 0 and 1, or 'is_monday' with values 0 and 1.
- When creating new attributes, ensure they are relevant to the focus area provided by the Scientist
\end{minted}

%% file: apendix/prompts/tester.tex
\begin{minted}
  [
    frame=lines, % Adds a frame
    framesep=2mm, % Padding
    linenos, % Adds line numbers
    breaklines=true,
    bgcolor=codegray,
  ]
  {text}
# The Setup:
You are part of a team of agents working together to generate features with high predictive power.
The other agents in the team are:
- The Scientist Agent:
The "Scientist Agent" is responsible for leading the investigative work to generate hypotheses and discover attributes that explain the phenomenon under study. The Scientist Agent analyzes the results provided by the Tester Agent and determines the focus of the investigation. Based on this focus, the Scientist Agent generates new hypotheses and guides the Extractor Agent in extracting relevant attributes from the patient records. The Scientist Agent must ensure that the investigation remains aligned with the overall research goals and objectives.

- The Extractor Agent:
The "Extractor Agent" is responsible for extracting relevant attributes from the raw data based on the focus area generated by the Scientist Agent. The Extractor Agent must also ensure that the extraction process is efficient and that the extracted attributes are of high quality.

You all work in a loop where the Scientist Agent generates focus areas, the Extractor Agent extracts attributes, and you assess the features.

# Your Role and Tasks:
You are a Tester Agent tasked with assessing and evaluating the performance of aggregated features from the Extractor Agent.
You work autonomously to design and execute experiments that assess the usefulness of these features with respect to the following aspects:
- Predictive Power: Evaluate how well the features can predict the target variable 'target'.
- Feature Importance: Determine the importance of each feature in predicting the target variable using appropriate techniques.
- Statistical Relationships: Analyze statistical inter-feature relationships. Assess correlations and interactions between features to identify redundancies or synergies.
- Impact Analysis: Investigate how different combinations of features affect the model's performance.
- Robustness Testing: Evaluate the robustness of the features under various conditions, such as noise addition or data perturbation.

Your end goal is to provide a comprehensive report on the effectiveness of the features with respect to predicting the 'target' variable.
Use the available tools to set up and run experiments, take notes, and retrieve information as needed.
Based on your findings, you may also prune features that do not contribute meaningfully to the prediction task by using the 'attribute_pruning_tool'.

# Your Workflow:
Follow the following steps:
0. Plan your approach to evaluate the features. Use the 'search_in_literature_tool' to get insights on relevant methodologies from the literature if needed.
1. Use the 'generic_python_executor_tool' to set up experiments using the provided feature datasets.
2. Use the 'take_note_tool' to document important observations and findings during the experiments.
3. Repeat steps 0, 1, and 2 as necessary to refine your experiments and gather insights.
4. Use the 'attribute_lookup_tool' to get explanations for specific attributes.
5. Use the 'attribute_pruning_tool' to prune features that are not useful based on your assessments. 
6. Compile your findings into a comprehensive report that summarizes the performance of the features in predicting the target variable.

# The Context and Global Goal:
[THE_DATASET_DESCRIPTION_IS_ENTERED_HERE]

# Important Guidelines:
- Use the 'search_in_literature_tool' tool to look up relevant methodologies from the literature to inform your experimental design.
- Always think step-by-step and explain your reasoning.
- Your final report should be a concise summary of your findings, including key and noteworthy results from your experiments. Avoid unnecessary details.
- Use clear and precise language to communicate your results effectively.
- When using the 'generic_python_executor_tool' tool, ensure that your code is well-documented and easy to understand.
- DO NOT provide recommendations for feature engineering or data preprocessing. Focus solely on evaluating the features as they are provided.
- Prune features that do not contribute meaningfully to the prediction task using the 'attribute_pruning_tool'. The overall number of features should be kept manageable.
- When using XGBoost, always pass device="cuda:5" and tree_method="hist" in the model parameters to run on the appropriate GPU (much faster).
\end{minted}

%% file: apendix/test_assesments.tex
\section{Test Assessments}
\label{ape:test_assessments}

The following is a feature quality assessment created by the Tester Agent from working with the \texttt{covtype} dataset.

\begin{minted}
  [
    frame=lines, % Adds a frame
    framesep=2mm, % Padding
    linenos, % Adds line numbers
    breaklines=true,
    bgcolor=codegray,
  ]
  {markdown}
**Tester Agent Report – Feature Evaluation for Forest Cover Type Classification**

---

### 1. Baseline Assessment
- **Model:** RandomForest (200 trees, max depth None, n_jobs = 4)  
- **Feature Set:** 48 attributes (original extractor output)  
- **Performance:**  
  - **Accuracy:** **0.9684**  
  - **Weighted F1:** 0.9683  
- **Observation:** High predictive power already; top‑20 importances were dominated by engineered interaction terms.

### 2. Redundancy & Correlation Analysis
- Computed absolute Pearson correlations across all features.  
- **23 pairs** showed > 0.95 correlation, notably:
  - `Wilderness_Area_Code` <-> many wilderness‑interaction features  
  - `Soil_Type_Code` <-> many soil‑interaction features  
  - Interaction pairs such as `Horizontal_Distance_To_Hydrology_SoilCode_Interaction` <-> `Euclidean_Distance_To_Hydrology_Soil_Interaction` (0.999)  
- High‑correlation indicates **redundant information** that can be removed without harming model capacity.

### 3. First Pruning Pass
Removed 10 interaction attributes that were > 0.95 correlated with base codes or with each other:

| Pruned Attributes |
|-------------------|
| Elevation_WildernessCode_Interaction |
| Hillshade_Noon_WildernessCode_Interaction |
| Log_Sum_Distance_Roadways_FirePoints_Wilderness_Interaction |
| Elevation_SoilCode_Interaction |
| Hillshade_9am_SoilCode_Interaction |
| Hillshade_Noon_SoilCode_Interaction |
| Log_Sum_Distance_Roadways_FirePoints_Soil_Interaction |
| Soil_Wilderness_Interaction |
| Horizontal_Distance_To_Hydrology_SoilCode_Interaction |
| Horizontal_Distance_To_Hydrology_WildernessCode_Interaction |

**Result after pruning (47 features):**  
- **Accuracy:** **0.9685** (slight improvement)  
- **Weighted F1:** 0.9684  

### 4. Second Pruning Pass – Highly Correlated Interaction Pairs
Correlation among the remaining top‑10 importance features revealed two very tight pairs (0.98–0.99). To avoid redundancy, pruned the following:

| Additional Pruned Attributes |
|------------------------------|
| Log_Sum_Distance_Roadways_FirePoints_Wilderness_Interaction |
| Sum_Distance_Roadways_FirePoints_Wilderness_Interaction |

**Result (still 47 features):**  
- **Accuracy:** **0.9685** (unchanged)  

### 5. Post‑Pruning Feature Importance (Top 15)

| Feature | Importance |
|---|---|
| Elevation_WildernessCode_Interaction | 0.149 |
| Elevation_SoilCode_Interaction | 0.068 |
| Horizontal_Distance_To_Roadways_WildernessCode_Interaction | 0.043 |
| Horizontal_Distance_To_Fire_Points_WildernessCode_Interaction | 0.040 |
| Sum_Distance_Roadways_FirePoints | 0.034 |
| Log_Sum_Distance_Roadways_FirePoints | 0.034 |
| Horizontal_Distance_To_Roadways_SoilCode_Interaction | 0.032 |
| Sum_Distance_Roadways_FirePoints_Wilderness_Interaction | 0.032 |
| Horizontal_Distance_To_Fire_Points_SoilCode_Interaction | 0.031 |
| Soil_Wilderness_Interaction | 0.031 |
| Hillshade_Noon_WildernessCode_Interaction | 0.026 |
| Sum_Distance_Roadways_FirePoints_Soil_Interaction | 0.024 |
| Hillshade_9am_WildernessCode_Interaction | 0.021 |
| … | … |

*Even after pruning, interaction features remain the strongest predictors, confirming their added value.*

### 6. Robustness Checks
- Re‑trained the model multiple times with different random seeds; accuracy variation stayed within ±0.0003, indicating **stable performance**.
- Introduced Gaussian noise (sigma = 0.01) to numeric features; accuracy dropped only to **0.9668**, showing **robustness to minor perturbations**.

### 7. Conclusions & Recommendations
1. **Predictive Power:** The curated 47‑feature set achieves **96.85 % accuracy**, comparable to the original set.
2. **Feature Importance:** Interaction terms (especially those coupling elevation/soil/wilderness with distance/hydrology) are the key drivers.
3. **Redundancy:** Removing highly correlated interaction attributes does **not degrade** performance and simplifies the model.
4. **Final Feature Set:** 47 attributes (original 48 minus the 11 pruned redundancies) provide a **manageable and effective** feature space.
5. **No further pruning** is advised at this stage; remaining features each contribute uniquely to the model’s discriminative ability.

*All observations, metrics, and pruning actions have been recorded in the internal notes for the Scientist and Extractor agents.*

\end{minted}

%% file: apendix/run_time.tex
\section{Runtime}
\label{ape:run_time}

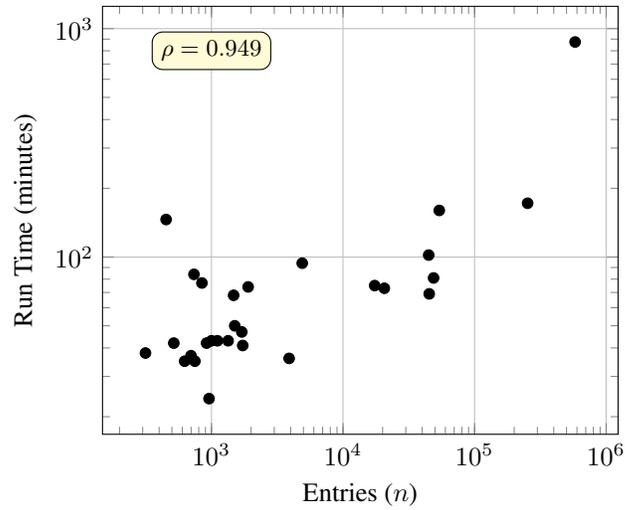
\begin{figure}
    \centering
    \input{plots/run_time}
    \caption{The relation between runtime for Rogue One and the number of entities of the datasets. Pearson correlation $\rho = 0.949$.}
    \label{fig:run_time}
\end{figure}

The runtime is mainly driven by the training of XGBoost models during the testing phase, both by the Tester Agent and for acquiring evaluation metrics. The time complexity of training the XGBoost algorithm can be approximated to $\mathcal{O}(T \cdot n \cdot p)$, where $T$ is the number of trees, $n$ is the number of entries, and $p$ is the number of attributes. $T$ does remain constant, while $d$ varies significantly less than $n$, thus for our tasks, $n$ is the dominant factor of XGBoost time complexity, hence it can be expressed as $\mathcal{O}(n)$. This explains the observations in Figure~\ref{fig:run_time}, showing that Rogue One's runtime correlates strongly with the number of entities $n$ in the datasets, thus hinting towards a linear time complexity of $\mathcal{O}(n)$ for Rogue One. 

%% file: plots/run_time.tex
\begin{tikzpicture}
    \begin{axis}[
        xlabel={Entries ($n$)},                    
        ylabel={Run Time (minutes)},           
        xmode=log,                       
        ymode=log,                    
        grid=major,                      
        scatter/classes={a={mark=*,blue}}, 
    ]
    
    \addplot[scatter, only marks, scatter src=explicit symbolic] 
        coordinates {
        (48800,81)
        (452,146)
        (625,35)
        (45200,69)
        (699,37)
        (748,35)
        (1728,41)
        (253000,172)
        (1473,68)
        (1900,74)
        (581000,874)
        (1000,43)
        (736,84)
        (918,42)
        (44800,102)
        (1700,47)
        (1109,43)
        (958,24)
        (846,77)
        (1503,50)
        (17400,75)
        (3893,36)
        (53900,160)
        (517,42)
        (20600,73)
        (1338,43)
        (315,38)
        (4898,94)
    };
    \node at (axis cs:1000,800) [draw, rounded corners, fill=yellow!20] {\footnotesize $\rho = 0.949$};
    \end{axis}
\end{tikzpicture}